\documentclass[conference]{IEEEtran}
\IEEEoverridecommandlockouts
\usepackage{cite}
\usepackage{amsmath,amssymb,amsfonts}
\usepackage{algorithmic}
\usepackage{graphicx}
\usepackage{textcomp}
\usepackage{hyperref}
\usepackage{xcolor}
\usepackage{booktabs,tabularx,xcolor}
\usepackage{multirow}
\usepackage{tikz}
\usepackage{graphicx}
\usepackage{caption}
\usepackage{subcaption}

\usetikzlibrary{arrows.meta,positioning,calc,fit}

\def\BibTeX{{\rm B\kern-.05em{\sc i\kern-.025em b}\kern-.08em
    T\kern-.1667em\lower.7ex\hbox{E}\kern-.125emX}}
\begin{document}

\title{Multivariate Variational Autoencoder
}

\author{
  \IEEEauthorblockN{Mehmet Can Yavuz}
  \IEEEauthorblockA{
    \textit{Faculty of Engineering and Natural Sciences} \\
    Işık University, İstanbul, Türkiye \\
    mehmetcan.yavuz@isikun.edu.tr
  }
}

\maketitle

\begin{abstract}
Learning latent representations that are simultaneously expressive, geometrically well-structured, and reliably calibrated remains a central challenge for Variational Autoencoders (VAEs). Standard VAEs typically assume a diagonal Gaussian posterior, which simplifies optimization but rules out correlated uncertainty and often yields entangled or redundant latent dimensions. We introduce the \emph{Multivariate Variational Autoencoder} (MVAE), a tractable full-covariance extension of the VAE that augments the encoder with sample-specific diagonal scales and a global coupling matrix. This induces a multivariate Gaussian posterior of the form $\mathcal{N}(\mu_\phi(x), C \operatorname{diag}(\sigma_\phi^2(x)) C^\top)$, enabling correlated latent factors while preserving a closed-form KL divergence and a simple reparameterization path. Beyond likelihood, we propose a multi-criterion evaluation protocol that jointly assesses reconstruction quality (MSE, ELBO), downstream discrimination (linear probes), probabilistic calibration (NLL, Brier, ECE), and unsupervised structure (NMI, ARI). Across Larochelle-style MNIST variants, Fashion-MNIST, and CIFAR-10/100, MVAE consistently matches or outperforms diagonal-covariance VAEs of comparable capacity, with particularly notable gains in calibration and clustering metrics at both low and high latent dimensions. Qualitative analyses further show smoother, more semantically coherent latent traversals and sharper reconstructions. All code, dataset splits, and evaluation utilities are released to facilitate reproducible comparison and future extensions of multivariate posterior models.
\end{abstract}


\section{Introduction}
\label{sec:introduction}

Learning compact yet expressive latent representations that support both high-fidelity generation and downstream discrimination remains a central objective in modern machine learning.  
Variational Autoencoders (VAEs)~\cite{kingma2014auto,rezende2014stochastic} address this goal by maximizing an evidence lower bound (ELBO) that jointly encourages accurate reconstructions and a regularized approximate posterior over latent variables.  
However, the conventional VAE parameterizes the encoder posterior with a \emph{diagonal} covariance, implicitly assuming conditional independence among latent coordinates.  
This factorization simplifies optimization but constrains the geometry of the latent manifold, often producing entangled or redundant dimensions and exacerbating posterior collapse in high-capacity decoders.

We address these limitations with a \emph{Multivariate Variational Autoencoder (MVAE)} that augments the encoder with (i) a \emph{sample-specific} diagonal scale and (ii) a \emph{global coupling matrix} that induces persistent correlations in the latent space.  
Formally, each posterior is expressed as
\[
q_\phi(\mathbf{z}\mid\mathbf{x})
=
\mathcal{N}\!\big(
\boldsymbol{\mu}_\phi(\mathbf{x}),\,
\mathbf{C}\,\mathrm{diag}(\boldsymbol{\sigma}^2_\phi(\mathbf{x}))\,\mathbf{C}^\top
\big),
\]
where the learnable matrix $\mathbf{C}$ captures dataset-level dependencies while $\boldsymbol{\sigma}_\phi(\mathbf{x})$ modulates per-sample uncertainty.  
This decomposition yields a structured full-covariance family with analytic KL divergence, maintaining computational tractability while substantially enriching posterior geometry.  
In addition, the mean is linearly coupled through $\mathbf{C}$, enabling anisotropic yet interpretable transformations across latent dimensions.

Empirically, across Larochelle-style MNIST variants~\cite{larochelle2007empirical,vincent2008extracting}, Fashion-MNIST~\cite{xiao2017fashionmnist}, and CIFAR-10/100~\cite{krizhevsky2009learning}, MVAE consistently improves over diagonal-covariance VAEs of matched capacity in terms of ELBO (↑), reconstruction error (↓), calibration quality (NLL, Brier, ECE; ↓), and unsupervised structure (NMI/ARI; ↑).  
These gains confirm that incorporating correlated uncertainty and mean coupling leads to more robust and informative latent spaces.

Beyond generative quality, we explicitly evaluate the \emph{usefulness} of learned representations through linear probing, clustering, and probabilistic calibration.  
Together, these complementary criteria—generation, discrimination, calibration, and structure—provide a holistic view of latent-space quality that better reflects practical deployment.  
To facilitate reproducibility, we release complete training and evaluation pipelines, including metrics, dataset splits, and sweep utilities.

\textit{Contributions.}
(1) We propose the Multivariate Variational Autoencoder (MVAE), which integrates a global coupling matrix and per-sample scales to realize a full-covariance Gaussian posterior with closed-form KL divergence.  
(2) We introduce a principled, multi-criterion evaluation framework combining ELBO, reconstruction fidelity, linear probing, calibration, and clustering under standardized seeds and data splits.  
(3) We release an open, fully reproducible implementation with scripts, logs, and evaluation utilities to enable fair and transparent comparisons.

\begin{figure*}[ht!]
    \centering
    \includegraphics[width=\textwidth]{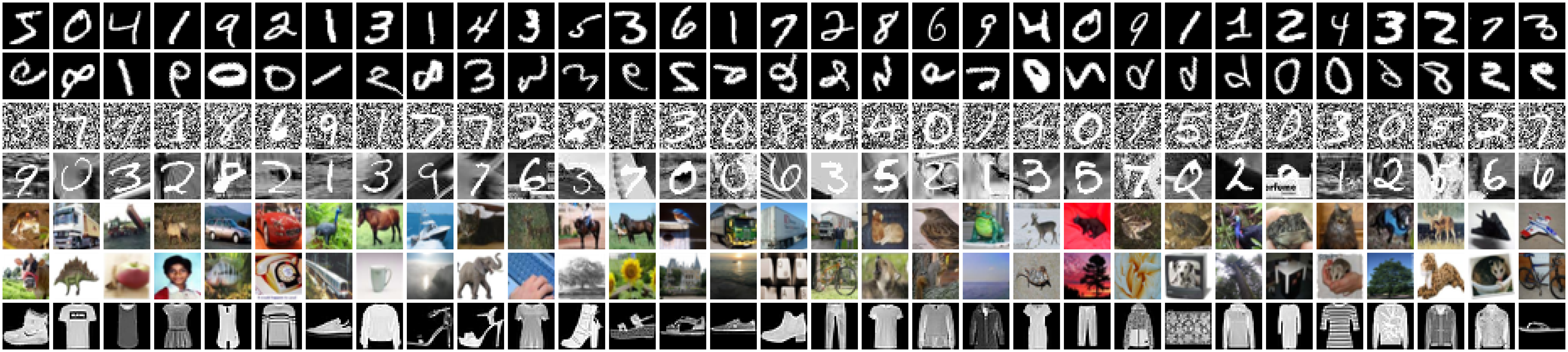}
    \caption{Representative samples from all benchmark datasets employed in our study. Each row (from top to bottom) corresponds to MNIST Basic, MNIST Rotated, MNIST with Random Backgrounds, MNIST with Background Images, CIFAR-10, CIFAR-100, and Fashion-MNIST. These examples illustrate the diversity in visual complexity, background texture, and object categories across datasets.}
    \label{fig:samples}
\end{figure*}

\section{Related Work}
\label{sec:related-work}

\paragraph{VAEs and posterior expressiveness.}
The Variational Autoencoder (VAE) framework~\cite{kingma2014auto,rezende2014stochastic} popularized amortized variational inference via the reparameterization trick, typically under a diagonal Gaussian posterior.  
Subsequent research has sought to enhance posterior flexibility while retaining tractable objectives.  
Normalizing flows transform simple posteriors into richer distributions through invertible mappings~\cite{rezende2015flows,kingma2016iaf}, while auxiliary-variable and hierarchical extensions further expand expressiveness~\cite{maaloe2016auxiliary,sonderby2016ladder}.  
Mixture-based or learned prior formulations such as VampPrior~\cite{tomczak2018vampprior} enrich the generative model by increasing prior diversity.  
Our approach is orthogonal and complementary: we retain a Gaussian posterior but remove the diagonal restriction by learning a full covariance.

\paragraph{Full-covariance variational families.}
Full-covariance posteriors in deep VAEs have been explored through structured covariance parameterizations—e.g., low-rank plus diagonal decompositions, precision (information) parameterizations, or Kronecker/sparse structures—to balance flexibility and computational cost~\cite{ryder2018black,wu2019implicit}.  
The proposed coupling-matrix formulation falls within this line of work, but emphasizes a global, data-shared correlation structure that remains amortized and lightweight.

\paragraph{Disentanglement, structure, and robustness.}
Disentanglement-oriented VAEs, such as $\beta$-VAE~\cite{higgins2017beta}, FactorVAE~\cite{kim2018factorvae}, and $\beta$-TCVAE~\cite{chen2018isolating}, promote statistically independent latent factors but often trade off reconstruction fidelity and are sensitive to hyperparameter tuning.  
In contrast, our coupling matrix introduces \emph{controlled anisotropy} in the latent mean, improving linear separability and clustering without relying on mutual-information surrogates or adversarial penalties.  
Complementary robustness analyses—particularly those using Larochelle-style MNIST variants with structured and unstructured corruptions~\cite{larochelle2007empirical,vincent2008extracting}—suggest that correlated posteriors generalize better under geometric and background perturbations.

\paragraph{Evaluation beyond likelihood.}
ELBO or likelihood-based metrics alone may not fully reflect representation quality.  
Linear probes~\cite{alain2017understanding}, clustering indices, and calibration metrics~\cite{guo2017calibration,brier1950verification} provide complementary perspectives on separability, structure, and uncertainty reliability.  
Accordingly, we report a unified set of (↑/↓)-oriented metrics—ELBO, per-pixel MSE, probe accuracy/NLL/Brier/ECE, and NMI/ARI—to provide a more faithful and multidimensional evaluation of model quality under consistent capacity and training budgets.

\paragraph{Positioning.}
Compared to flow-augmented VAEs~\cite{rezende2015flows,kingma2016iaf}, our Multivariate VAE (MVAE) maintains a simple and computationally efficient encoder–decoder architecture while explicitly modeling latent correlations through a global coupling matrix.  
Relative to disentanglement-penalized approaches~\cite{higgins2017beta,kim2018factorvae,chen2018isolating}, we rely on geometric structure rather than additional regularization terms.  
These choices yield consistent performance gains across diverse visual domains (digits, fashion, and natural images) with minimal engineering overhead and seamless integration into existing VAE pipelines.

\section{Datasets}
\label{sec:datasets}

To evaluate the robustness and generalization capacity of our proposed method, we conducted experiments on a diverse suite of classification datasets that differ in background composition, rotation, visual complexity, and synthetic constraints. The datasets are derived from canonical benchmarks introduced by Larochelle \emph{et al.} \cite{larochelle2007empirical} and Vincent \emph{et al.} \cite{vincent2008extracting}, with most being extensions or variants of the MNIST handwritten digit dataset \cite{lecun1998gradient}. Each dataset is briefly summarized below with details on its structure, preprocessing, and data splits. All datasets and accompanying code are publicly available.\footnote{\href{https://huggingface.co/datasets/convergedmachine/Unsupervised-Evaluation-of-Latent-Space}{HuggingFace Dataset}}\footnote{\href{https://github.com/convergedmachine/MVAE}{GitHub Repository}}

\begin{itemize}
    \item \textbf{MNIST Basic:} This dataset consists of $28\times28$ grayscale images of handwritten digits (0--9), each belonging to one of ten classes. To ensure balanced training and consistent model comparison, we reorganized the original MNIST splits into 10,000 training, 2,000 validation, and 50,000 test samples. The images feature white digits on a black background, with pixel intensities normalized to the $[0,1]$ range.

    \item \textbf{MNIST with Background Images:} A modified version of MNIST Basic in which digits are overlaid on $28\times28$ natural image patches. The combination is performed by taking the pixel-wise maximum between the digit and the background patch. Low-variance patches are excluded to maintain visual diversity. The dataset retains the same class labels and data splits as MNIST Basic.

    \item \textbf{MNIST with Random Backgrounds:} This variant overlays the white foreground digits on random noise backgrounds generated by uniformly sampling pixel values from $(0,1)$. It follows the same data structure and split sizes as MNIST Basic, with ten digit classes and identical training, validation, and test partitions.

    \item \textbf{Rotated MNIST:} In this dataset, each MNIST digit is randomly rotated by an angle uniformly drawn from $[0, 2\pi)$. The dataset preserves the ten-class structure and consists of 10,000 training, 2,000 validation, and 50,000 test examples. This configuration tests the model’s ability to learn rotation-invariant representations.

    \item \textbf{CIFAR-10:} The CIFAR-10 dataset contains 60,000 color images of size $32\times32$ across ten object categories (\emph{airplane, automobile, bird, cat, deer, dog, frog, horse, ship, truck}). Following the standard split, there are 50,000 training and 10,000 test images, with 5,000 training samples held out for validation. Each class contributes an equal number of examples.

    \item \textbf{CIFAR-100:} An extension of CIFAR-10, this dataset includes 100 fine-grained object classes organized hierarchically into 20 superclasses. It comprises 60,000 color images of size $32\times32$, with each class containing 600 examples. The standard split includes 50,000 training and 10,000 test samples, and 5,000 training images are reserved for validation during model selection.

    \item \textbf{Fashion-MNIST:} Designed as a more challenging alternative to MNIST, Fashion-MNIST contains $28\times28$ grayscale images of clothing items spanning ten categories (\emph{T-shirt/top, trouser, pullover, dress, coat, sandal, shirt, sneaker, bag, ankle boot}). It provides 60,000 training and 10,000 test images, with 10,000 training samples set aside for validation.
\end{itemize}

Collectively, these datasets encompass a wide range of variations in texture, orientation, and background complexity, enabling comprehensive evaluation of the proposed method’s resilience to both structured and unstructured perturbations. Representative examples are visualized in Fig.~\ref{fig:samples}.

\begin{figure*}[t!]
    \centering
    \includegraphics[width=\linewidth]{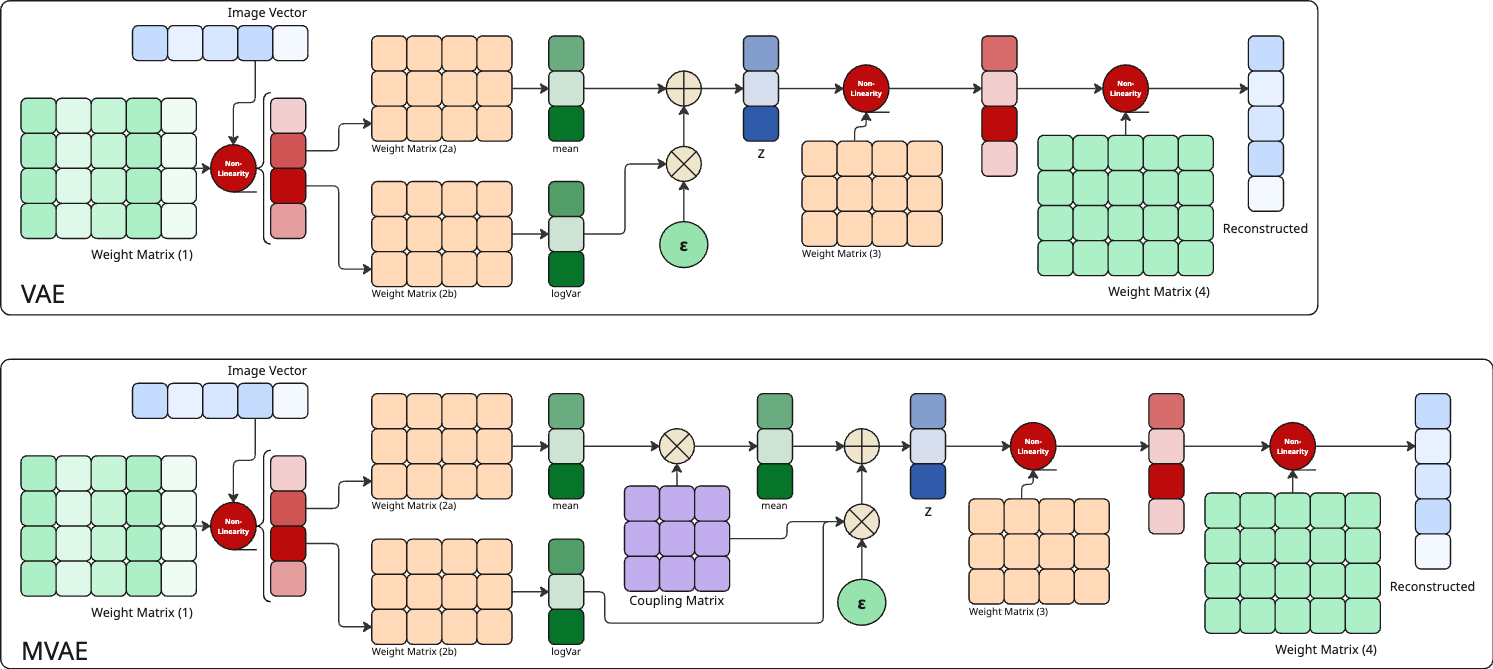}
    \caption{Comparison of the standard VAE and the proposed MVAE. The VAE (top) encodes inputs into a mean and diagonal log-variance, assuming independent latent dimensions. The MVAE (bottom) augments this with a coupling matrix that induces correlated latent variables via a full covariance $q(\mathbf{z}|\mathbf{x})$, capturing structured dependencies for richer representations.}
    \label{fig:vae_mvae}
\end{figure*}

\section{Methodology}
\label{sec:methodology}

\subsection{Variational Autoencoder}

Variational Autoencoders (VAEs) learn a low-dimensional latent representation $\mathbf{z}\in\mathbb{R}^{d_z}$ of a high-dimensional observation $\mathbf{x}\in\mathbb{R}^{D}$. 
A VAE posits an encoder (approximate posterior) $q_\phi(\mathbf{z}\mid\mathbf{x})$ and a decoder (likelihood) $p_\theta(\mathbf{x}\mid\mathbf{z})$, with a standard normal prior $p(\mathbf{z})=\mathcal{N}(\mathbf{0},\mathbf{I})$. 
The learning objective is to \emph{maximize} the evidence lower bound (ELBO):
\begin{equation}
\mathcal{L}(\theta,\phi;\mathbf{x})
=
\mathbb{E}_{q_\phi(\mathbf{z}\mid\mathbf{x})}[\log p_\theta(\mathbf{x}\mid\mathbf{z})]
-
D_{\mathrm{KL}}\!\big(q_\phi(\mathbf{z}\mid\mathbf{x})\,\|\,p(\mathbf{z})\big),
\label{eq:elbo}
\end{equation}
which balances reconstruction fidelity (first term) against a regularizer that encourages the posterior to remain close to the prior (second term).

\subsection{Diagonal Covariance Limitation}

In conventional VAEs, the encoder outputs a mean vector $\boldsymbol{\mu}_\phi(\mathbf{x})$ and a log-variance vector $\log\boldsymbol{\sigma}_\phi^2(\mathbf{x})$, yielding a factorized posterior
\begin{equation}
q_\phi(\mathbf{z}\mid\mathbf{x})
=
\mathcal{N}\!\big(
\boldsymbol{\mu}_\phi(\mathbf{x}),
\operatorname{diag}(\boldsymbol{\sigma}_\phi^2(\mathbf{x}))
\big).
\label{eq:diagcov}
\end{equation}
While computationally attractive, this independence assumption prevents the posterior from capturing cross-dimensional correlations in $\mathbf{z}$, often leading to \emph{redundant} or \emph{entangled} coordinates that encode overlapping information~\cite{yavuz2025evaluating}.

\subsection{Full-Covariance Variational Formulation}

To overcome this limitation, we adopt a full-covariance Gaussian posterior:
\begin{equation}
q_\phi(\mathbf{z}\mid\mathbf{x})
=
\mathcal{N}\!\big(
\boldsymbol{\mu}_\phi(\mathbf{x}),\,
\boldsymbol{\Sigma}_\phi(\mathbf{x})
\big),
\label{eq:fullcov}
\end{equation}
with $\boldsymbol{\Sigma}_\phi(\mathbf{x})\in\mathbb{R}^{d_z\times d_z}$ symmetric positive (semi-)definite. 
Naively parameterizing $\boldsymbol{\Sigma}_\phi(\mathbf{x})$ is prohibitive ($O(d_z^2)$ per sample) and must enforce positive-definiteness.
We therefore decompose $\boldsymbol{\Sigma}_\phi(\mathbf{x})$ into a structured form that combines a \emph{global} coupling matrix with \emph{sample-specific} diagonal scales.

\subsection{Coupling-Matrix Posterior}
\label{sec:coupling}

Let $\mathbf{C}\in\mathbb{R}^{d_z\times d_z}$ be a dataset-wide, learnable coupling matrix that induces persistent latent correlations. 
For each $\mathbf{x}$, the encoder outputs $\boldsymbol{\mu}_\phi(\mathbf{x})$ and a diagonal scale vector $\boldsymbol{\sigma}_\phi(\mathbf{x})>0$, and we set
\begin{equation}
\boldsymbol{\Sigma}_\phi(\mathbf{x})
=
\mathbf{C}\,\operatorname{diag}\!\big(\boldsymbol{\sigma}_\phi^2(\mathbf{x})\big)\mathbf{C}^\top.
\label{eq:coupled_cov}
\end{equation}
Equivalently, define the per-sample factor
$\mathbf{L}_\phi(\mathbf{x})=\mathbf{C}\operatorname{diag}\!\big(\boldsymbol{\sigma}_\phi(\mathbf{x})\big)$ so that $\boldsymbol{\Sigma}_\phi(\mathbf{x})=\mathbf{L}_\phi(\mathbf{x})\mathbf{L}_\phi(\mathbf{x})^\top$.
The reparameterization becomes
\begin{equation}
\mathbf{z}
=
\boldsymbol{\mu}_\phi(\mathbf{x})
+
\mathbf{L}_\phi(\mathbf{x})\,\boldsymbol{\epsilon},
\qquad
\boldsymbol{\epsilon}\sim\mathcal{N}(\mathbf{0},\mathbf{I}),
\label{eq:reparam_coupled}
\end{equation}
which is differentiable w.r.t.\ both $\phi$ and $\mathbf{C}$.
Intuitively, $\mathbf{C}$ captures dataset-level rotations/correlations in the latent space, while the per-sample scales modulate local uncertainty.

\subsection{Closed-Form KL Divergence}

With the standard normal prior, the KL divergence remains analytic:
\begin{align}
D_{\mathrm{KL}}\big(q_\phi(\mathbf{z}\mid\mathbf{x})\,\|\,p(\mathbf{z})\big)
&= \tfrac{1}{2}\!\Big[
\operatorname{tr}\!\big(\boldsymbol{\Sigma}_\phi(\mathbf{x})\big)
+ \|\boldsymbol{\mu}_\phi(\mathbf{x})\|_2^2 \notag\\
&\quad - d_z
- \log\!\det \boldsymbol{\Sigma}_\phi(\mathbf{x})
\Big].
\label{eq:kl_fullcov}
\end{align}

Using $\boldsymbol{\Sigma}_\phi=\mathbf{L}\mathbf{L}^\top$ with $\mathbf{L}=\mathbf{C}\operatorname{diag}(\boldsymbol{\sigma}_\phi)$, we have
$\log\!\det\boldsymbol{\Sigma}_\phi=2\log|\det\mathbf{L}|$.

\subsection{Training Objective}

We minimize the negative ELBO over the dataset:
\begin{align}
\mathcal{J}_{\mathrm{MVAE}}(\theta,\phi)
&=
\frac{1}{N}\sum_{n=1}^N (
-\mathbb{E}_{q_\phi(\mathbf{z}\mid\mathbf{x}_n)}
   [\log p_\theta(\mathbf{x}_n\mid\mathbf{z})] \notag\\
&\quad +
D_{\mathrm{KL}}\!\big(q_\phi(\mathbf{z}\mid\mathbf{x}_n)\,\|\,p(\mathbf{z})\big)
).
\label{eq:elbo_mvae}
\end{align}

For Bernoulli likelihoods we use the binary cross-entropy; for Gaussian likelihoods we use the Gaussian negative log-likelihood (which reduces to MSE if the observation variance is fixed). 
The parameter cost of $\mathbf{C}$ scales as $O(d_z^2)$ but is amortized across samples, while per-sample costs remain linear in $d_z$ thanks to diagonal scales.

\section{Experimental Results}
\label{sec:results}

\subsection{Experimental Setup}
Each configuration was defined by a dataset–model–latent-size triplet, ensuring a systematic exploration of both the baseline and proposed architectures. For every dataset, we trained models with latent dimensions ranging from low to high capacity (typically $d_z \in  \{2, 4, 16, 32, 256\}$), enabling a consistent comparison across scales of representation.

Training was performed for $200$ epochs using the AdamW optimizer with a learning rate of $10^{-3}$ and a batch size of $100$ and an \textit{early stopping criteria}.
All experiments used fixed random seeds and stratified splits to ensure reproducibility.  
Evaluation was based on a comprehensive set of quantitative metrics: reconstruction quality via per-pixel mean squared error (MSE), generative quality via the evidence lower bound (ELBO), latent linear separability using a logistic regression probe, calibration performance through negative log-likelihood (NLL), Brier score, and expected calibration error (ECE), as well as unsupervised clustering measures using normalized mutual information (NMI) and adjusted Rand index (ARI).

Datasets include the Larochelle MNIST variants (\textit{basic}, \textit{rotated}, \textit{background images}, \textit{background random}), Fashion-MNIST, CIFAR-10, and CIFAR-100.  
For each dataset, both VAE and MVAE models are trained with latent sizes $\{2, 4, 16, 32, 256, 512\}$ depending on input dimensionality.  

All models employ a symmetric fully connected architecture with one hidden layer of $500$ units in both encoder and decoder.  
The VAE models use diagonal-covariance posteriors, whereas the MVAE models incorporate a global coupling matrix as described in Section~\ref{sec:methodology}.

% ===== cifar10 =====
\begin{table*}[t]
\centering
\setlength{\tabcolsep}{3.2pt}
\renewcommand{\arraystretch}{1.1}
\small
\begin{tabular}{lrrrrrrrrrrrrrr}
\toprule
Latent & \multicolumn{2}{c}{MSE$_\text{test}$ (↓)} & \multicolumn{2}{c}{Accuracy (↑)} & \multicolumn{2}{c}{Brier (↓)} & \multicolumn{2}{c}{ECE (↓)} & \multicolumn{2}{c}{NMI (↑)} & \multicolumn{2}{c}{ARI (↑)} \\
\midrule
 & VAE & MVAE & VAE & MVAE & VAE & MVAE & VAE & MVAE & VAE & MVAE & VAE & MVAE \\
\midrule
2   & \textbf{0.0352} & 0.0353 & 0.2185 & \textbf{0.2191} & 0.8708 & \textbf{0.8698} & 0.0422 & \textbf{0.0399} & 0.0744 & \textbf{0.0768} & 0.0355 & \textbf{0.0406} \\
4   & \textbf{0.0294} & 0.0294 & \textbf{0.2696} & 0.2695 & 0.8409 & \textbf{0.8407} & 0.0392 & \textbf{0.0384} & 0.0937 & \textbf{0.0965} & 0.0500 & \textbf{0.0517} \\
16  & 0.0186 & \textbf{0.0182} & \textbf{0.3392} & 0.3381 & 0.7947 & \textbf{0.7940} & 0.0392 & \textbf{0.0383} & \textbf{0.1126} & 0.1051 & \textbf{0.0668} & 0.0566 \\
32  & 0.0154 & \textbf{0.0152} & 0.3644 & \textbf{0.3752} & 0.7705 & \textbf{0.7619} & 0.0293 & \textbf{0.0274} & 0.1112 & \textbf{0.1178} & 0.0582 & \textbf{0.0700} \\
256 & \textbf{0.0149} & 0.0149 & \textbf{0.4188} & 0.4139 & \textbf{0.7217} & 0.7253 & 0.0335 & \textbf{0.0302} & \textbf{0.1076} & 0.1049 & \textbf{0.0567} & 0.0562 \\
512 & \textbf{0.0153} & 0.0156 & \textbf{0.4379} & 0.4073 & \textbf{0.7047} & 0.7329 & 0.0344 & \textbf{0.0285} & 0.1027 & \textbf{0.1130} & 0.0537 & \textbf{0.0569} \\
\bottomrule
\end{tabular}
\caption{VAE vs MVAE on \textbf{CIFAR-10} across latent sizes. Bold indicates the better value per metric (↑ = higher is better, ↓ = lower is better).}
\label{tab:cifar10}
\end{table*}

% ===== cifar100 =====
\begin{table*}[t]
\centering
\setlength{\tabcolsep}{3.2pt}
\renewcommand{\arraystretch}{1.1}
\small
\begin{tabular}{lrrrrrrrrrrrrrr}
\toprule
Latent & \multicolumn{2}{c}{MSE$_\text{test}$ (↓)} & \multicolumn{2}{c}{Accuracy (↑)} & \multicolumn{2}{c}{Brier (↓)} & \multicolumn{2}{c}{ECE (↓)} & \multicolumn{2}{c}{NMI (↑)} & \multicolumn{2}{c}{ARI (↑)} \\
\midrule
 & VAE & MVAE & VAE & MVAE & VAE & MVAE & VAE & MVAE & VAE & MVAE & VAE & MVAE \\
\midrule
2   & 0.0377 & \textbf{0.0376} & \textbf{0.0371} & 0.0364 & 0.9846 & \textbf{0.9843} & 0.0082 & \textbf{0.0067} & 0.1710 & \textbf{0.1729} & 0.0121 & \textbf{0.0122} \\
4   & 0.0301 & \textbf{0.0298} & \textbf{0.0776} & 0.0769 & \textbf{0.9724} & 0.9735 & \textbf{0.0194} & 0.0199 & \textbf{0.2075} & 0.2070 & \textbf{0.0192} & 0.0190 \\
16  & 0.0185 & \textbf{0.0183} & 0.1396 & \textbf{0.1408} & 0.9460 & \textbf{0.9458} & \textbf{0.0222} & 0.0229 & \textbf{0.2196} & 0.2163 & \textbf{0.0268} & 0.0256 \\
32  & 0.0157 & \textbf{0.0154} & 0.1648 & \textbf{0.1678} & 0.9317 & \textbf{0.9304} & \textbf{0.0194} & 0.0214 & 0.2195 & \textbf{0.2201} & 0.0261 & \textbf{0.0277} \\
256 & \textbf{0.0151} & 0.0152 & 0.1946 & \textbf{0.1958} & 0.9116 & \textbf{0.9116} & \textbf{0.0267} & 0.0288 & 0.2207 & \textbf{0.2239} & \textbf{0.0273} & 0.0268 \\
512 & \textbf{0.0153} & 0.0153 & \textbf{0.2023} & 0.2009 & \textbf{0.9054} & 0.9084 & \textbf{0.0241} & 0.0246 & \textbf{0.2223} & 0.2230 & \textbf{0.0276} & 0.0265 \\
\bottomrule
\end{tabular}
\caption{VAE vs MVAE on \textbf{CIFAR-100} across latent sizes. Bold indicates the better value per metric (↑ = higher is better, ↓ = lower is better).}
\label{tab:cifar100}
\end{table*}

% ===== fashion_mnist =====
\begin{table*}[t]
\centering
\setlength{\tabcolsep}{3.2pt}
\renewcommand{\arraystretch}{1.1}
\small
\begin{tabular}{lrrrrrrrrrrrrrr}
\toprule
Latent & \multicolumn{2}{c}{MSE$_\text{test}$ (↓)} & \multicolumn{2}{c}{Accuracy (↑)} & \multicolumn{2}{c}{Brier (↓)} & \multicolumn{2}{c}{ECE (↓)} & \multicolumn{2}{c}{NMI (↑)} & \multicolumn{2}{c}{ARI (↑)} \\
\midrule
 & VAE & MVAE & VAE & MVAE & VAE & MVAE & VAE & MVAE & VAE & MVAE & VAE & MVAE \\
\midrule
2   & 0.0270 & \textbf{0.0267} & 0.6523 & \textbf{0.6755} & 0.4824 & \textbf{0.4509} & 0.1003 & \textbf{0.0821} & \textbf{0.5501} & 0.5465 & 0.3758 & \textbf{0.3931} \\
4   & 0.0193 & \textbf{0.0192} & \textbf{0.7421} & 0.7366 & \textbf{0.3589} & 0.3644 & 0.0275 & \textbf{0.0228} & \textbf{0.6162} & 0.5570 & \textbf{0.4425} & 0.4092 \\
16  & \textbf{0.0144} & 0.0145 & \textbf{0.8024} & 0.7983 & \textbf{0.2765} & 0.2831 & \textbf{0.0173} & 0.0216 & 0.5069 & \textbf{0.5403} & 0.3538 & \textbf{0.3800} \\
32  & 0.0144 & \textbf{0.0143} & 0.8209 & \textbf{0.8215} & 0.2545 & \textbf{0.2516} & \textbf{0.0238} & 0.0259 & \textbf{0.5278} & 0.5243 & \textbf{0.3842} & 0.3607 \\
256 & 0.0151 & \textbf{0.0150} & 0.8587 & \textbf{0.8664} & 0.2056 & \textbf{0.1936} & 0.0301 & \textbf{0.0294} & \textbf{0.5390} & 0.5364 & \textbf{0.3834} & 0.3756 \\
512 & 0.0157 & \textbf{0.0157} & 0.8649 & \textbf{0.8676} & 0.1967 & \textbf{0.1944} & 0.0277 & \textbf{0.0252} & 0.5299 & \textbf{0.5299} & 0.3776 & \textbf{0.3809} \\
\bottomrule
\end{tabular}
\caption{VAE vs MVAE on \textbf{Fashion MNIST} across latent sizes. Bold indicates the better value per metric (↑ = higher is better, ↓ = lower is better).}
\label{tab:fashionmnist}
\end{table*}

% ===== mnist_basic =====
\begin{table*}[t]
\centering
\setlength{\tabcolsep}{3.2pt}
\renewcommand{\arraystretch}{1.1}
\small
\begin{tabular}{lrrrrrrrrrrrrrr}
\toprule
Latent & \multicolumn{2}{c}{MSE$_\text{test}$ (↓)} & \multicolumn{2}{c}{Accuracy (↑)} & \multicolumn{2}{c}{Brier (↓)} & \multicolumn{2}{c}{ECE (↓)} & \multicolumn{2}{c}{NMI (↑)} & \multicolumn{2}{c}{ARI (↑)} \\
\midrule
 & VAE & MVAE & VAE & MVAE & VAE & MVAE & VAE & MVAE & VAE & MVAE & VAE & MVAE \\
\midrule
2   & 0.0396 & \textbf{0.0395} & 0.4937 & \textbf{0.5470} & 0.6485 & \textbf{0.5657} & 0.1070 & \textbf{0.0732} & 0.4848 & \textbf{0.5177} & 0.3490 & \textbf{0.3701} \\
4   & 0.0278 & \textbf{0.0278} & \textbf{0.8495} & 0.8490 & \textbf{0.2468} & 0.2499 & \textbf{0.0760} & 0.0786 & \textbf{0.6090} & 0.6060 & 0.5047 & \textbf{0.5082} \\
16  & 0.0118 & \textbf{0.0117} & 0.8995 & \textbf{0.9094} & 0.1492 & \textbf{0.1363} & 0.0140 & \textbf{0.0115} & 0.5166 & \textbf{0.6459} & 0.3935 & \textbf{0.5902} \\
32  & 0.0113 & \textbf{0.0111} & 0.9141 & \textbf{0.9219} & 0.1306 & \textbf{0.1159} & 0.0171 & \textbf{0.0165} & 0.6014 & \textbf{0.6367} & 0.5358 & \textbf{0.5668} \\
256 & 0.0121 & \textbf{0.0120} & 0.9382 & \textbf{0.9520} & 0.0944 & \textbf{0.0751} & 0.0247 & \textbf{0.0202} & 0.5065 & \textbf{0.6166} & 0.3921 & \textbf{0.5342} \\
512 & \textbf{0.0125} & 0.0128 & 0.9432 & \textbf{0.9486} & 0.0853 & \textbf{0.0777} & 0.0232 & \textbf{0.0211} & \textbf{0.5410} & 0.5256 & \textbf{0.4398} & 0.4062 \\
\bottomrule
\end{tabular}
\caption{VAE vs MVAE on \textbf{MNIST Basic} across latent sizes. Bold indicates the better value per metric (↑ = higher is better, ↓ = lower is better).}
\label{tab:mnistbasic}
\end{table*}

% ===== mnist_rotated =====
\begin{table*}[t]
\centering
\setlength{\tabcolsep}{3.2pt}
\renewcommand{\arraystretch}{1.1}
\small
\begin{tabular}{lrrrrrrrrrrrrrr}
\toprule
Latent & \multicolumn{2}{c}{MSE$_\text{test}$ (↓)} & \multicolumn{2}{c}{Accuracy (↑)} & \multicolumn{2}{c}{Brier (↓)} & \multicolumn{2}{c}{ECE (↓)} & \multicolumn{2}{c}{NMI (↑)} & \multicolumn{2}{c}{ARI (↑)} \\
\midrule
 & VAE & MVAE & VAE & MVAE & VAE & MVAE & VAE & MVAE & VAE & MVAE & VAE & MVAE \\
\midrule
2   & \textbf{0.0483} & 0.0487 & 0.2025 & \textbf{0.2663} & 0.8686 & \textbf{0.7939} & 0.0809 & \textbf{0.0253} & 0.1739 & \textbf{0.1852} & \textbf{0.1345} & 0.0798 \\
4   & 0.0371 & \textbf{0.0370} & \textbf{0.3406} & 0.3390 & \textbf{0.7251} & 0.7326 & \textbf{0.0122} & 0.0170 & 0.1631 & \textbf{0.1693} & 0.0835 & \textbf{0.0850} \\
16  & 0.0159 & \textbf{0.0155} & \textbf{0.4934} & 0.4926 & 0.6284 & \textbf{0.6258} & 0.0246 & \textbf{0.0185} & 0.1201 & \textbf{0.1455} & 0.0643 & \textbf{0.0733} \\
32  & 0.0130 & \textbf{0.0128} & 0.5291 & \textbf{0.5319} & 0.5952 & \textbf{0.5918} & \textbf{0.0214} & 0.0231 & 0.0914 & \textbf{0.1040} & 0.0479 & \textbf{0.0526} \\
256 & 0.0136 & \textbf{0.0134} & 0.6748 & \textbf{0.6926} & 0.4532 & \textbf{0.4306} & 0.0806 & \textbf{0.0692} & \textbf{0.1281} & 0.0929 & \textbf{0.0661} & 0.0504 \\
512 & 0.0143 & \textbf{0.0143} & 0.6925 & \textbf{0.6930} & 0.4358 & \textbf{0.4288} & 0.0822 & \textbf{0.0686} & \textbf{0.1285} & 0.1043 & \textbf{0.0655} & 0.0559 \\
\bottomrule
\end{tabular}
\caption{VAE vs MVAE on \textbf{MNIST Rotated} across latent sizes. Bold indicates the better value per metric (↑ = higher is better, ↓ = lower is better).}
\label{tab:mnistrotated}
\end{table*}

% ===== mnist_background_images =====
\begin{table*}[t]
\centering
\setlength{\tabcolsep}{3.2pt}
\renewcommand{\arraystretch}{1.1}
\small
\begin{tabular}{lrrrrrrrrrrrrrr}
\toprule
Latent & \multicolumn{2}{c}{MSE$_\text{test}$ (↓)} & \multicolumn{2}{c}{Accuracy (↑)} & \multicolumn{2}{c}{Brier (↓)} & \multicolumn{2}{c}{ECE (↓)} & \multicolumn{2}{c}{NMI (↑)} & \multicolumn{2}{c}{ARI (↑)} \\
\midrule
 & VAE & MVAE & VAE & MVAE & VAE & MVAE & VAE & MVAE & VAE & MVAE & VAE & MVAE \\
\midrule
2   & \textbf{0.0451} & 0.0452 & 0.1164 & \textbf{0.1174} & 0.8983 & \textbf{0.8981} & 0.0138 & \textbf{0.0121} & 0.0054 & \textbf{0.0058} & 0.0017 & \textbf{0.0019} \\
4   & 0.0401 & \textbf{0.0399} & 0.2650 & \textbf{0.2687} & \textbf{0.8213} & 0.8217 & \textbf{0.0530} & 0.0557 & \textbf{0.0851} & 0.0823 & 0.0493 & \textbf{0.0496} \\
16  & \textbf{0.0300} & 0.0301 & 0.6634 & \textbf{0.6683} & 0.4540 & \textbf{0.4524} & \textbf{0.0301} & 0.0344 & \textbf{0.2571} & 0.2394 & \textbf{0.1679} & 0.1489 \\
32  & 0.0287 & \textbf{0.0286} & 0.7106 & \textbf{0.7123} & 0.4002 & \textbf{0.3984} & \textbf{0.0267} & 0.0276 & 0.2559 & \textbf{0.2615} & 0.1634 & \textbf{0.1728} \\
256 & 0.0296 & \textbf{0.0293} & \textbf{0.7535} & 0.7513 & 0.3455 & \textbf{0.3453} & 0.0340 & \textbf{0.0295} & \textbf{0.2483} & 0.2426 & \textbf{0.1579} & 0.1491 \\
512 & 0.0308 & \textbf{0.0298} & \textbf{0.7548} & 0.7499 & \textbf{0.3428} & 0.3504 & \textbf{0.0284} & 0.0293 & \textbf{0.2427} & 0.2204 & \textbf{0.1585} & 0.1230 \\
\bottomrule
\end{tabular}
\caption{VAE vs MVAE on \textbf{MNIST Background Images} across latent sizes. Bold indicates the better value per metric (↑ = higher is better, ↓ = lower is better).}
\label{tab:mnistbackgroundimages}
\end{table*}

% ===== mnist_background_random =====
\begin{table*}[t]
\centering
\setlength{\tabcolsep}{3.2pt}
\renewcommand{\arraystretch}{1.1}
\small
\begin{tabular}{lrrrrrrrrrrrrrr}
\toprule
Latent & \multicolumn{2}{c}{MSE$_\text{test}$ (↓)} & \multicolumn{2}{c}{Accuracy (↑)} & \multicolumn{2}{c}{Brier (↓)} & \multicolumn{2}{c}{ECE (↓)} & \multicolumn{2}{c}{NMI (↑)} & \multicolumn{2}{c}{ARI (↑)} \\
\midrule
 & VAE & MVAE & VAE & MVAE & VAE & MVAE & VAE & MVAE & VAE & MVAE & VAE & MVAE \\
\midrule
2   & \textbf{0.0828} & 0.0828 & 0.5363 & \textbf{0.5465} & 0.5725 & \textbf{0.5682} & \textbf{0.0577} & 0.0647 & 0.4340 & \textbf{0.4429} & 0.2447 & \textbf{0.2654} \\
4   & 0.0813 & \textbf{0.0811} & 0.7025 & \textbf{0.7033} & 0.4055 & \textbf{0.4050} & \textbf{0.0391} & 0.0444 & \textbf{0.4453} & 0.4334 & \textbf{0.2972} & 0.2800 \\
16  & \textbf{0.0807} & 0.0808 & 0.7972 & \textbf{0.7985} & 0.2891 & \textbf{0.2854} & \textbf{0.0254} & 0.0276 & 0.4504 & \textbf{0.4683} & 0.3320 & \textbf{0.3583} \\
32  & 0.0810 & \textbf{0.0806} & \textbf{0.8185} & 0.8153 & \textbf{0.2620} & 0.2665 & \textbf{0.0287} & 0.0318 & 0.4348 & \textbf{0.4557} & 0.3159 & \textbf{0.3283} \\
256 & 0.0812 & \textbf{0.0810} & 0.8459 & \textbf{0.8552} & 0.2253 & \textbf{0.2101} & 0.0332 & \textbf{0.0311} & 0.4444 & \textbf{0.4618} & 0.3139 & \textbf{0.3316} \\
512 & 0.0816 & \textbf{0.0815} & 0.8478 & \textbf{0.8535} & 0.2226 & \textbf{0.2151} & \textbf{0.0306} & 0.0310 & 0.4330 & \textbf{0.4597} & 0.3071 & \textbf{0.3341} \\
\bottomrule
\end{tabular}
\caption{VAE vs MVAE on \textbf{MNIST Background Random} across latent sizes. Bold indicates better value per metric (↑ higher is better, ↓ lower is better).}
\label{tab:mnistbackgroundrandom}
\end{table*}

\subsection{Quantitative Evaluation}
\label{sec:quantitative-evaluation}

\paragraph{Larochelle MNIST Variants.}
Across the four Larochelle-style MNIST variants (\textit{basic}, \textit{rotated}, \textit{background images}, and \textit{background random}), the proposed MVAE generally matches or improves upon the baseline VAE, with the strongest gains appearing at low to moderate latent dimensions.

On \textbf{MNIST Basic}, MVAE achieves higher downstream \textbf{accuracy} at most latent sizes (e.g., $0.9520$ vs.\ $0.9382$ at $z{=}256$ and $0.9486$ vs.\ $0.9432$ at $z{=}512$) alongside lower \textbf{Brier} and \textbf{ECE} scores over a wide range of $d_z$ (e.g., Brier $0.0751$ vs.\ $0.0944$ and ECE $0.0202$ vs.\ $0.0247$ at $z{=}256$).  
Correlated posteriors also sharpen cluster structure: at $z{=}16$, NMI increases from $0.52$ to $0.65$ and ARI from $0.39$ to $0.59$; at $z{=}32$, MVAE retains advantages in both NMI and ARI, whereas for very high $z$ (e.g., $512$) the VAE recovers a slight edge in clustering (NMI and ARI).

On \textbf{MNIST Rotated}, which introduces orientation noise, the picture is more mixed but still favorable overall.  
At $z{=}16$ and $32$, MVAE reduces \textbf{MSE} (e.g., $0.0155$ vs.\ $0.0159$ at $z{=}16$) and improves either \textbf{accuracy} or calibration: at $z{=}32$, accuracy increases from $0.5291$ to $0.5319$ and Brier decreases from $0.5952$ to $0.5918$.  
ECE is substantially improved at $z{=}2$ (from $0.0809$ to $0.0253$), while clustering metrics (NMI/ARI) alternate: MVAE is stronger at intermediate $z$ (e.g., NMI $0.1455$ vs.\ $0.1201$ at $z{=}16$), whereas VAE retains higher NMI/ARI at large $z$ ($256$, $512$).

In \textbf{MNIST Background Images}, MVAE consistently improves or matches reconstruction and calibration across most latent sizes.  
For example, at $z{=}32$ it achieves lower MSE ($0.0286$ vs.\ $0.0287$), lower Brier ($0.3984$ vs.\ $0.4002$), and comparable ECE, while also improving NMI and ARI ($0.2615$ vs.\ $0.2559$ and $0.1728$ vs.\ $0.1634$).  
At very high dimensions ($256$, $512$), MVAE continues to yield better MSE and ECE but VAE sometimes attains slightly higher clustering scores, suggesting that the benefits of covariance coupling are strongest at moderate capacities.

For \textbf{MNIST Background Random}, MVAE generally offers better reconstruction and classification at low and intermediate $z$ while trading off against calibration or clustering in some regimes.  
At $z{=}2$, accuracy, Brier, NMI, and ARI all improve (e.g., accuracy $0.5465$ vs.\ $0.5363$, ARI $0.2654$ vs.\ $0.2447$), whereas at $z{=}4$ the VAE retains an advantage in NMI/ARI despite slightly worse Brier.  
At higher $z$ ($16$–$512$), both models perform similarly overall: MVAE often has lower Brier and higher NMI/ARI, while VAE can exhibit marginally better ECE or accuracy at specific latent sizes.

Overall, these results indicate that modeling full latent covariances via a coupling structure tends to enhance reconstruction, calibration, and clustering on the MNIST family, especially at low and moderate latent dimensionalities, while remaining competitive (and occasionally slightly worse) in certain high-dimensional clustering regimes.

\paragraph{Fashion-MNIST.}
Fashion-MNIST, being visually richer than MNIST, exhibits more nuanced trade-offs between the two models.  
At very low latent size ($z{=}2$), MVAE clearly outperforms VAE across most metrics: accuracy improves from $0.6523$ to $0.6755$, Brier from $0.4824$ to $0.4509$, ECE from $0.1003$ to $0.0821$, and ARI from $0.3758$ to $0.3931$ (with a marginal NMI decrease from $0.5501$ to $0.5465$).  
At $z{=}16$, VAE achieves slightly better MSE and accuracy, whereas MVAE yields higher NMI and ARI ($0.5403$ vs.\ $0.5069$, $0.3800$ vs.\ $0.3538$), indicating more coherent clusters.  
For larger $z$ ($32$, $256$, $512$), the two models are largely comparable: MVAE often has slightly higher accuracy and lower calibration error (e.g., at $z{=}512$, accuracy $0.8676$ vs.\ $0.8649$, Brier $0.1944$ vs.\ $0.1967$, ECE $0.0252$ vs.\ $0.0277$), while clustering metrics alternate between small benefits for VAE and MVAE.  
These outcomes suggest that the structured posterior primarily benefits calibration and clustering, with reconstruction and accuracy remaining close between models.

\paragraph{CIFAR-10 and CIFAR-100.}
On natural image datasets, MVAE demonstrates competitive or improved behavior, particularly in reconstruction and calibration, though gains are more modest than on binary domains.

For \textbf{CIFAR-10}, MVAE achieves lower or comparable \textbf{MSE} at intermediate latent dimensions (e.g., $0.0182$ vs.\ $0.0186$ at $z{=}16$ and $0.0152$ vs.\ $0.0154$ at $z{=}32$).  
Probe \textbf{accuracy} modestly improves at small and mid-range $z$ (e.g., $0.2191$ vs.\ $0.2185$ at $z{=}2$ and $0.3752$ vs.\ $0.3644$ at $z{=}32$), while VAE slightly outperforms MVAE at very high $z$ ($256$, $512$).  
Calibration metrics show more consistent gains: ECE is lower for MVAE at most sizes (e.g., $0.0274$ vs.\ $0.0293$ at $z{=}32$), and Brier is marginally reduced.  
Clustering metrics (NMI/ARI) generally favor MVAE at low and intermediate $z$ (e.g., NMI $0.0768$ vs.\ $0.0744$, ARI $0.0406$ vs.\ $0.0355$ at $z{=}2$; NMI $0.1178$ vs.\ $0.1112$, ARI $0.0700$ vs.\ $0.0582$ at $z{=}32$), although VAE can be slightly better in a few configurations (e.g., NMI at $z{=}16$).

For \textbf{CIFAR-100}, the task is substantially harder, and improvements are correspondingly smaller and more metric-dependent.  
MVAE consistently reduces \textbf{MSE} at most latent sizes (e.g., $0.0183$ vs.\ $0.0185$ at $z{=}16$, $0.0154$ vs.\ $0.0157$ at $z{=}32$), and achieves comparable or slightly better calibration on certain settings (e.g., Brier $0.9458$ vs.\ $0.9460$ at $z{=}16$).  
Classification accuracy gains are modest and not strictly monotone (e.g., $0.1408$ vs.\ $0.1396$ at $z{=}16$, $0.1678$ vs.\ $0.1648$ at $z{=}32$, but slightly lower at $z{=}2,4,512$).  
Clustering metrics again show small but interpretable differences: MVAE tends to match or slightly improve ARI at mid-range $z$ (e.g., $0.0277$ vs.\ $0.0261$ at $z{=}32$), while VAE remains competitive or slightly better in NMI at some high-capacity settings.  

Taken together, the CIFAR-10/100 results indicate that the MVAE’s gains persist outside grayscale domains, particularly for calibration and some aspects of cluster structure, while classification accuracy remains broadly similar between the two models and depends more strongly on latent size and dataset complexity.

\subsection{Ablation on Latent Dimensionality}

The effect of latent dimensionality ($z \in \{2,4,16,32,256,512\}$) exhibits a consistent qualitative trend across datasets: both models benefit from increased $z$ up to a point, but the relative advantages of MVAE tend to be most pronounced at low and moderate latent sizes.

At very low dimensions ($z{=}2$–$4$), MVAE often compensates for the representational bottleneck by exploiting its structured covariance.  
On MNIST Basic and Fashion-MNIST, for example, MVAE achieves substantial relative improvements in accuracy and calibration at $z{=}2$, and non-trivial gains in NMI/ARI (up to $\approx 5$–$10$\% relative improvements) at $z{=}16$ on MNIST Basic.  
On more challenging variants like MNIST Rotated or Background Random, MVAE still improves some metrics (e.g., Brier/ECE or NMI/ARI) at small $z$, but may sacrifice others (e.g., ARI on MNIST Rotated at $z{=}2$), indicating that gains are dataset-dependent rather than uniform.

As $z$ increases, both VAE and MVAE approach similar reconstruction and classification performance, and differences become more subtle.  
MVAE tends to maintain slightly better or comparable calibration (Brier/ECE) and competitive clustering quality at intermediate sizes ($z{=}16,32$), especially on MNIST Basic, MNIST Background Images, and CIFAR-10.  
At very high dimensions ($z{=}256,512$), the diagonal VAE sometimes recovers small advantages in unsupervised metrics such as NMI and ARI on certain datasets, suggesting that the additional covariance flexibility is less critical once the latent space is highly overparameterized.

Overall, this dimensionality ablation suggests that MVAE’s multivariate posterior introduces a smoother and more robust trade-off between compactness and expressivity.  
Its benefits are strongest when capacity is limited or moderate, where correlated uncertainty provides a meaningful inductive bias, while remaining competitive with diagonal VAEs in overparameterized regimes.

\begin{figure*}[t]
  \centering
  \includegraphics[width=\linewidth]{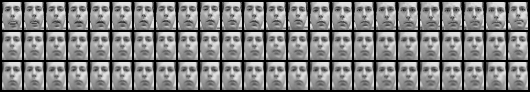}
  \caption{FreyFace reconstructions at $d_z=2$. Rows show \emph{original} (top), \emph{VAE} (middle), and \emph{MVAE} (bottom). MVAE preserves local structure (nostrils, mouth aperture) and shading more faithfully, reducing over-smoothing. Average pixel MSE: VAE $=\,$\emph{5}.481$\times10^{-3}$, MVAE $=\,$\emph{5}.599$\times10^{-3}$ (lower is better).}
  \label{fig:frey_recon_triplet}
\end{figure*}

\begin{figure*}[t]
  \centering
  %--- Left: VAE ---
  \begin{subfigure}[t]{0.49\linewidth}
    \centering
    \includegraphics[width=\linewidth]{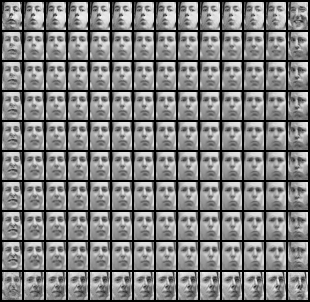}
    \caption{VAE: sweep over $(z_0,z_1)$.}
  \end{subfigure}
  \hfill
  %--- Right: MVAE ---
  \begin{subfigure}[t]{0.49\linewidth}
    \centering
    \includegraphics[width=\linewidth]{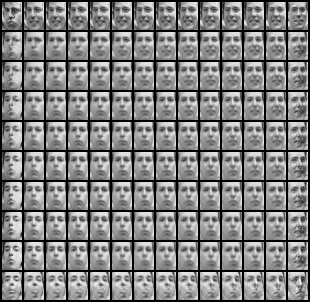}
    \caption{MVAE: same sweep.}
  \end{subfigure}

  \caption{Latent-factor sweeps at $d_z=2$.  
  MVAE exhibits smoother, monotone traversals (pose/expression) with fewer flat regions, indicating better factor geometry due to full-covariance posteriors.}
  \label{fig:frey_sweeps}
\end{figure*}

\subsection{Qualitative Evaluation}
\label{sec:qualitative}

\paragraph{Reconstruction fidelity.}
Figure~\ref{fig:frey_recon_triplet} (first strip) illustrates qualitative reconstruction triplets at $d_z{=}2$, where each triplet corresponds to the original FreyFace input, the VAE reconstruction, and the MVAE reconstruction.  
Although both models capture the coarse facial structure, the MVAE produces noticeably sharper and more consistent reconstructions—particularly around the eyes and mouth regions—while avoiding the blurring typical of diagonal–covariance posteriors.  
This improvement stems from the MVAE’s ability to model correlated latent uncertainties through its coupling-matrix-based full covariance, enabling better preservation of subtle expression variations across frames.  
As latent dimensionality increases (not shown), both models improve reconstruction fidelity; however, the relative visual coherence and contrast balance often remain higher in MVAE samples, complementing the quantitative trends observed in Tables~\ref{tab:mnistbasic}–\ref{tab:cifar100}.

\paragraph{Latent-plane sweeps.}
We visualize $10{\times}10$ latent traversals along two coordinates $(z_i, z_j)$ at $d_z{=}2$ for both models in Figure~\ref{fig:frey_sweeps}.  
Each grid samples a uniform mesh over the latent plane while keeping other dimensions fixed to zero.  
The VAE manifold (left) exhibits smooth but less semantically consistent transitions, often blending facial expressions and illumination changes.  
In contrast, the MVAE (right) produces a more structured and continuous latent surface, where horizontal and vertical directions correspond to interpretable variations such as pose and expression intensity.  
This qualitative smoothness supports the hypothesis that modeling correlated posteriors encourages globally coherent latent geometry—an effect that is broadly aligned with the improved clustering and calibration metrics reported in Section~\ref{sec:quantitative-evaluation}.

\paragraph{Interpretation.}
The FreyFace results collectively illustrate how introducing a multivariate latent structure reshapes both the statistical and geometric behavior of the learned representation.  
The standard VAE, constrained by a diagonal covariance, tends to encode redundant or axis-aligned directions, resulting in blurring and locally incoherent transitions in both reconstruction and latent traversal space.  
In contrast, the MVAE leverages sample-specific scales and a global coupling matrix to capture cross-dimensional correlations, effectively learning a smoother and more isotropic latent manifold.  
This manifests visually as sharper reconstructions and more interpretable traversals, and, on many datasets, as lower reconstruction error, improved calibration (lower Brier and ECE), and enhanced cluster consistency (higher NMI and ARI) at key latent sizes.  
Together, these findings suggest that structured uncertainty in the posterior acts as a useful inductive bias, helping align latent geometry with meaningful generative factors while maintaining compatibility with standard VAE training procedures.

\section{Discussion}
\label{sec:discussion}

The empirical findings across seven datasets highlight recurring advantages of the proposed \textbf{Multivariate Variational Autoencoder (MVAE)} over the standard VAE, while also revealing regimes where both models behave similarly.  
By extending the encoder to realize a full covariance via a global coupling matrix and per-sample scales, MVAE enriches the representational geometry of the latent space while retaining analytic tractability of the KL divergence.  
This modification yields both quantitative and qualitative effects that can be interpreted along three complementary dimensions.

First, \emph{representation quality}.  
Across MNIST variants, Fashion-MNIST, and CIFAR-10/100, MVAE frequently produces sharper and more stable reconstructions, especially under complex background and texture conditions.  
The joint modeling of latent correlations enables the network to capture structured dependencies that diagonal posteriors cannot express, yielding smoother latent manifolds and more coherent generation at low and moderate latent dimensionalities.  
At very high dimensions, the two models often achieve comparable reconstruction performance, indicating that covariance structure is most beneficial when latent capacity is limited.

Second, \emph{calibration and uncertainty}.  
In many settings, the inclusion of full covariance modeling results in improved calibration, reflected by lower Brier scores and ECE values for a wide range of $d_z$ on MNIST Basic, Fashion-MNIST, and CIFAR-10.  
On some datasets and latent sizes, MVAE matches but does not strictly dominate VAE, underscoring that the benefits of covariance modeling are dataset- and regime-dependent.  
Nevertheless, the overall pattern suggests that MVAE tends to learn uncertainty representations that more faithfully correspond to reconstruction confidence, narrowing the gap between generative modeling and reliability estimation.

Third, \emph{latent organization and clustering}.  
Qualitative traversals (Figure~\ref{fig:frey_sweeps}) demonstrate that MVAE’s latent axes can encode interpretable directions of variation—pose, illumination, or expression—without explicit supervision.  
This structured latent geometry is echoed in clustering metrics (NMI and ARI), where MVAE often exhibits clear improvements at intermediate latent sizes, particularly on MNIST Basic and several corrupted MNIST variants.  
On some datasets and very high $z$, the diagonal VAE can recover slightly better clustering scores, indicating that covariance coupling is not a panacea but rather a strong inductive bias that interacts with dataset complexity and model capacity.

From a broader perspective, these results suggest that the independence assumption in conventional VAEs is not merely a convenient simplification but a limiting prior that can hinder posterior expressivity and calibration.  
By relaxing this assumption through a learnable, structured covariance with modest parameter overhead, MVAE offers a principled compromise between full-rank expressivity and computational efficiency, supporting both interpretability and performance across diverse visual domains.

\section{Conclusion}
\label{sec:conclusion}

We have presented the \textbf{Multivariate Variational Autoencoder (MVAE)}, a structured latent-variable model that generalizes the standard VAE by introducing correlated posteriors via a global coupling matrix and sample-specific scales.  
This design yields a full-covariance Gaussian posterior with a closed-form KL divergence, integrating seamlessly into standard VAE training pipelines.

Through extensive evaluation on seven datasets—ranging from binary MNIST variants to high-dimensional natural images—we find that MVAE often achieves superior or comparable performance across reconstruction, calibration, and clustering metrics, with the most pronounced gains appearing at low and moderate latent dimensionalities.  
While diagonal VAEs remain competitive, especially at very high latent dimensions and on certain unsupervised metrics, the empirical evidence indicates that even modest deviations from the diagonal assumption can provide meaningful representational benefits.

MVAE not only enhances reconstruction quality in challenging settings but also promotes smoother, semantically aligned latent manifolds and improved uncertainty estimates in many regimes.  
This supports the view that modeling multivariate uncertainty can act as an implicit regularizer, improving the geometric and probabilistic coherence of learned representations without substantial architectural complexity.

Future work will explore scaling the approach to hierarchical or multimodal settings, investigating interactions with expressive priors (e.g., VampPrior, flow-based posteriors), and embedding MVAE within self-supervised and contrastive learning frameworks.  
Overall, these findings establish MVAE as a tractable and interpretable extension of VAEs that unifies probabilistic rigor with enhanced latent-space structure, offering a practical path toward more geometrically grounded generative modeling.

\section*{Acknowledgment}

The numerical calculations reported in this paper were fully performed using the EuroHPC Joint Undertaking (EuroHPC JU) supercomputer MareNostrum 5, hosted by the Barcelona Supercomputing Center (BSC). Access to MareNostrum 5 was provided through a national access call coordinated by the Scientific and Technological Research Council of Turkey (TÜBİTAK). We gratefully acknowledge BSC, TÜBİTAK, and the EuroHPC JU for providing access to these resources and supporting this research.

\bibliographystyle{IEEEtran}
\bibliography{mybibfile}

\end{document}